\documentclass[runningheads]{llncs}
\usepackage{graphicx}
\usepackage{comment}
\usepackage{amsmath,amssymb} 
\usepackage{color}
\usepackage{multirow}
\usepackage{makecell}
\usepackage{threeparttable}
\usepackage{multicol}
\usepackage{subfigure}

\usepackage[colorlinks, citecolor=red, citecolor=green]{hyperref}
\usepackage{caption}
\begin{document}
\pagestyle{headings}
\mainmatter
\def\ECCVSubNumber{2014}  

\title{Conditional Sequential Modulation\\ for Efficient Global Image Retouching} 

\titlerunning{Conditional Sequential Modulation}
%
\author{Jingwen He\thanks{The first two authors are co-first authors. $\dagger$ Corresponding author}\inst{1,2} \and
Yihao Liu$^{\star}$\inst{1,2,3} \and
Yu Qiao\inst{1,2} \and Chao Dong$\dagger$\inst{1,2}}
\authorrunning{Jingwen He et al.}
%
\institute{ShenZhen Key Lab of Computer Vision and Pattern Recognition, SIAT-SenseTime Joint Lab, Shenzhen Institutes of Advanced Technology, Chinese Academy of Sciences\\ \and
SIAT Branch, Shenzhen Institute of Artificial Intelligence and Robotics for Society \and
University of Chinese Academy of Sciences \\
\email{\{jw.he, yh.liu4, yu.qiao, chao.dong\}@siat.ac.cn}}
\maketitle

\begin{abstract}
Photo retouching aims at enhancing the aesthetic visual quality of images that suffer from photographic defects such as over/under exposure, poor contrast, inharmonious saturation. Practically, photo retouching can be accomplished by a series of image processing operations. In this paper, we investigate some commonly-used retouching operations and mathematically find that these pixel-independent operations can be approximated or formulated by multi-layer perceptrons (MLPs). Based on this analysis, we propose an extremely light-weight framework - Conditional Sequential Retouching Network (CSRNet) - for efficient global image retouching. CSRNet consists of a base network and a condition network. The base network acts like an MLP that processes each pixel independently and the condition network extracts the global features of the input image to generate a condition vector. To realize retouching operations, we modulate the intermediate features using Global Feature Modulation (GFM), of which the parameters are transformed by condition vector. Benefiting from the utilization of $1\times1$ convolution, CSRNet only contains less than 37k trainable parameters, which is orders of magnitude smaller than existing learning-based methods. Extensive experiments show that our method achieves state-of-the-art performance on the benchmark MIT-Adobe FiveK dataset quantitively and qualitatively. Code is available at \url{https://github.com/hejingwenhejingwen/CSRNet}.
\end{abstract}

\begin{figure}
	\centering
	\includegraphics[height=4.2cm]{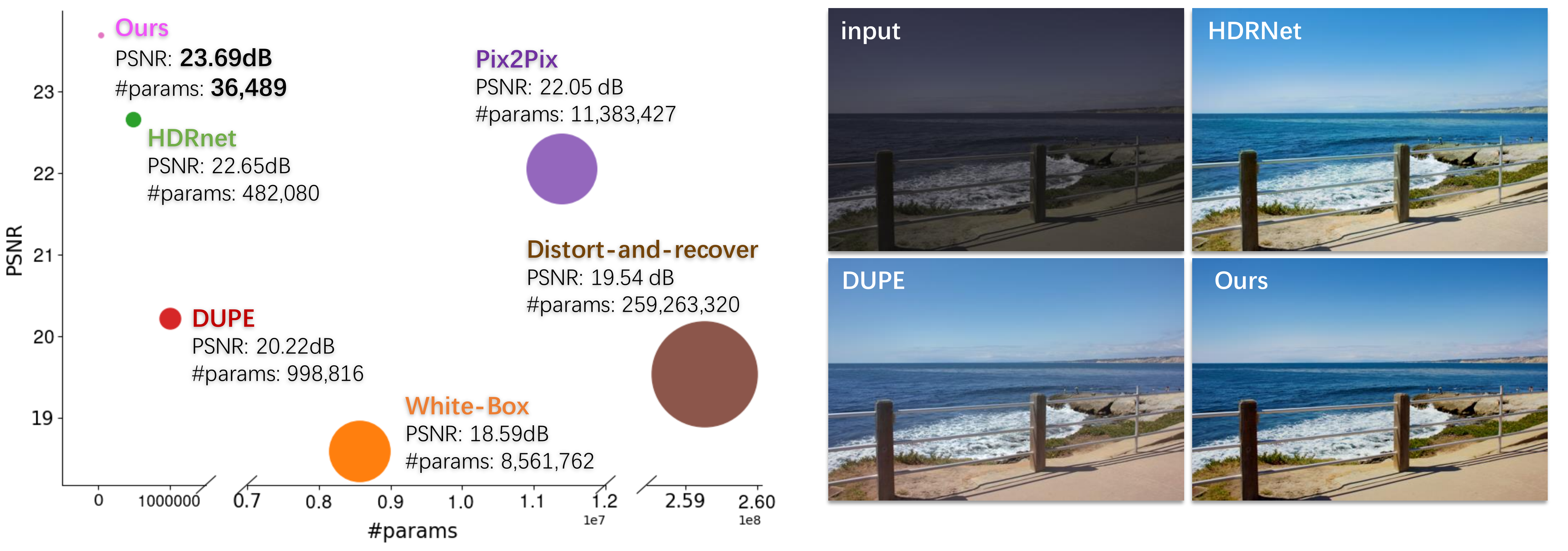}
	\caption{\textbf{Left:} Compared with existing state-of-the-art methods, our method achieves superior performance with extremely few parameters (1/13 of HDRNet \cite{hdrnet} and 1/250 of White-Box \cite{white-box}). The diameter of the circle represents the amount of trainable parameters. \textbf{Right:} Image retouching examples. Please zoom in for best view.}
	\label{fig:1}
\end{figure}

\section{Introduction}
Photo retouching can significantly improve the visual quality of photographs through a sequence of image processing operations, such as brightness and contrast changes. Manual retouching requires specialized skills and training, thus is challenging for causal users. Even for professional retouchers, dealing with large collections requires tedious repetitive editing works. This presents the needs for automatic photo retouching. It can be equipped in smart phones to help ordinary people get visual-pleasing photos, or it can be built in photo editing softwares to provide an editing reference for experts. 

The aim of photo retouching is to generate a high quality image from a low quality input. Recent learning-based methods tend to treat photo retouching as a special case of image enhancement or image-to-image translation. They use CNNs to learn either the transformation matrix \cite{hdrnet} or an end-to-end mapping \cite{DSLR,WESPE,DPE} from input/output pairs. Generally, photo retouching adjusts the global color tones, without the change of high frequency components (e.g., edges), while other image enhancement/translation tasks focus more on local patterns and will even change the image textures. Moreover, photo retouching is naturally a sequential processing, which can be decomposed into several independent simple operations. This property does not always hold for image enhancement and image-to-image translation problems. As most state-of-the-art algorithms \cite{hdrnet,Underexposed,DPE} are not specialized for photo retouching, they generally require extra parameters (e.g., $3\times3$ convolutions) to deal with local patterns, which could largely restrict their implementation efficiency. Detailed comparisons of different retouching methods are presented in the Related Work section. In real scenarios, most commonly-used operations (LUT, tone mapping, image enhancement operations in commercial software) are global adjustment. Thus in this paper, we focus on ``global" photo retouching without considering local operations.

To design an efficient photo retouching algorithm, we invesigate several retouching operations adopted in \cite{white-box,Distort} and find that these commonly-used operations (e.g., contrast adjustment, tone mapping) are location-independent/pixel-independent. The input pixels can be mapped to the output pixels via pixel-wise mapping functions, without the need of local image features. We take a step further and show that these pixel-wise functions can be approximated by multi-layer perceptrons (MLPs). Different adjustment operations can share similar network structures but with different parameters. Then the input image can be sequentially processed by a set of neural networks to generate the final output. 

Based on the above observation, we propose an extremely light-weight network - Conditional Sequential Retouching Network (CSRNet) - for fast global photo retouching. The key idea is to mimic the sequential processing procedure and implicitly model the editing operations in an end-to-end trainable network. The framework consists of two modules - the base network and the condition network. The base network adopts a fully convolutional structure. While the unique feature is that all filters are of size $1\times1$, indicating that each pixel is processed independently. Therefore, the base network can be regarded as an MLP for individual pixels. To realize retouching operations, we modulate the intermediate features using Global Feature Modulation (GFM), of which the parameters are controlled by the condition network. The condition network generates a condition vector, which is then broadcasted to different layers of the base network for feature modulation. This procedure is just like a sequential editing process operated on different stages of the MLP (see Figure \ref{fig:views}). These two modules are jointly optimized from human-adjusted image pairs. 

The proposed network enjoys a very simple architecture, which contains only six plain convolutional layers in total, without any complex building blocks. Such a compact network could achieve state-of-the-art performance on MIT-Adobe FiveK dataset \cite{mit-adobe}, with less than 37k parameters -1/13 of HDRNet \cite{hdrnet} and 1/90 of DPE \cite{DPE} (see Figure \ref{fig:1} and Table \ref{table:main}). We have also conducted extensive ablation studies on various settings, including trying different hand-crafted global priors, network structures and feature modulation strategies.

In addition to automatic adjustment, users will desire to control the output styles according to their own preference. Even for the same style, they may also want to adjust the overall retouching strength. To meet diverse user flavors, our method enjoys the flexibility to train different condition networks for different styles, without changing the base network. For the same style, we use image interpolation between input and output images to realize the strength control.

Our contributions are three-fold.
1. We propose the Conditional Sequential Retouching Network (CSRNet) for efficient global photo retouching. The proposed method can achieve state-of-the-art performance with less than 37k parameters. 
2. We combine the idea of color decomposition and sequential processing in a unified CNN framework, which could learn implicit step-wise retouching operations without intermediate supervision. 
3. We achieve continuous output effects among various retouching styles.  We find that image interpolation could realize strength control between different stylized images. 

\section{Related Work}
We briefly review the recent progress on image retouching and enhancement. Traditional algorithms have proposed various operations and filters to enhance the visual quality of images, such as histogram equalization, local Laplacian operator \cite{laplacian}, fast bilateral filtering \cite{Fastbilateral}, and color correction methods based on the gray-world \cite{Shades} or gray-edge \cite{Edge-based} assumption. Since Bychkovsky et al. \cite{mit-adobe} collected a large-scale dataset MIT-Adobe FiveK, which contains input and expert-retouched image pairs, a plenty of learning-based enhancing algorithms have been developed to continuously push the performance. Generally, these learning-based methods can be divided into three groups: physical-modeling-based methods, image-to-image translation methods and reinforcement learning methods. Physical-modeling-based methods attempt to estimate the intermediate parameters of the proposed physical models or assumptions for image enhancing. Based on the Retinex theory of color vision \cite{retinex}, several algorithms were developed for image exposure correction by estimating the reflectance and illumination with learnable models \cite{weightedvariational,cameraresponsemodel,correction,Underexposed}. By postulating that the enhanced output image can be expressed as local pointwise transformations of the input image, Gharbi et al. \cite{hdrnet} combined bilateral grid \cite{bilateralgrid} and bilateral guided upsampling models \cite{Bilateralguided}, then constructed a CNN model to predict the affine transformation coefficients in bilateral space for real-time image enhancement. Methods of the second group treat image enhancement as an image-to-image translation problem, which directly learn the end-to-end mapping between input and the enhanced image without modelling intermediate parameters. Ignatov et al. explored to translate ordinary photos into DSLR-quality images by residual convolutional neural networks \cite{DSLR} and weakly supervised generative adversarial networks \cite{WESPE}. Chen et al. \cite{DPE} utilized an improved two-way generative adversarial network (GAN) that can be trained in an unpair-learning manner. Reinforcement learning is adopted for image retouching, which aims at explicitly simulating the step-wise retouching process. Hu et al. \cite{white-box} presented a White-Box photo post-processing framework that learns to make decisions based on the current state of the image. Park et al. \cite{Distort} casted the color enhancement problem into a Markov Decision Process (MDP) where each action is defined as a global color adjustment operation and selected by Deep Q-Network \cite{deepQ}.

\section{Method}
Our method aims at fast automatic image retouching with low computation cost. First, we analyze several commonly-used retouching operations and gain important insights. Based on the analysis, we propose the framework -- Conditional Sequential Retouching Network (CSRNet). Then we illustrate the intrinsic working mechanism of CSRNet in two perspectives. Finally, we describe how to achieve different retouching styles and control the overall enhancement strength.

\begin{figure}[htbp]
	\centering
	\subfigure[An MLP on individual pixels.]{
		\begin{minipage}[t]{0.45\linewidth}
			\centering
			\includegraphics[width=2in]{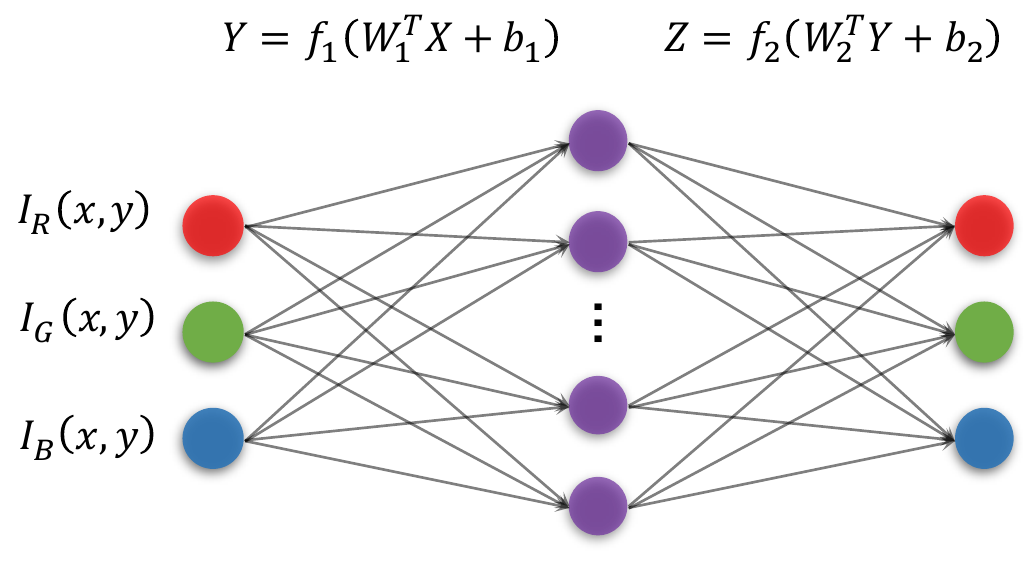}
		\end{minipage}
	}
	\subfigure[CSRNet. (k: kernel size; n: number of feature maps; s: stride.)]{
		\begin{minipage}[t]{0.45\linewidth}
			\centering
			\includegraphics[width=2in]{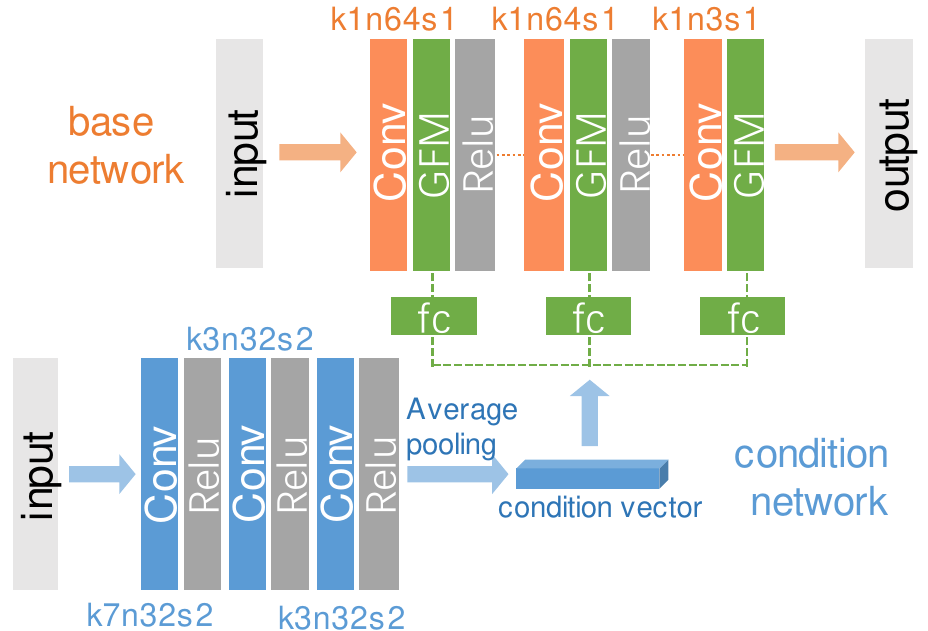}
		\end{minipage}
	}
	\caption{(a) Illustration for MLP on a single pixel. Pixel-independent operation can be viewed as an MLP on individual pixels, such as brightness change, white-balancing, saturation controlling and tone-mapping. (b) The proposed network consists of base network, condition network and GFM.
	}
\label{fig:mlp}
\end{figure}

\subsection{Analysis of Retouching Operations}
Image retouching is accomplished by a series of image processing operations, such as the manipulation of brightness/contrast, the adjustment in each color channel, and the controlling of saturation/hue/tones. We mathematically find that these pixel-independent operations can be approximated or formulated by multi-layer perceptrons (MLPs). Below we show two examples.\\
\textbf{Global brightness change.} Given an input image $I$, the global brightness is described as the average value of its luminance map: $I_Y=0.299*I_R+0.587*I_G+0.114*I_B$, where $I_R$, $I_G$, $I_B$ represent the RGB channels, respectively. One simple way to adjust the brightness is to multiply a scalar for each pixel:
\begin{equation}\label{con:brightness}
	I_{Y}^{'}(x,y)=\alpha I_Y(x,y) 
\end{equation}
where $I_{Y}^{'}(x,y)$ is the adjusted pixel value, $\alpha$ is the scalar, and $(x,y)$ indicates the pixel location in an $M \times N$ image. We can formulate the adjustment formula (\ref{con:brightness}) into the representation of an MLP:
\begin{equation}\label{mlp:brightness_mpl}
	Y=f(W^T X+b) 
\end{equation}
where $X\in\mathbb{R}^{MN}$ is the vector flattened from the input image, $W\in\mathbb{R}^{MN \times MN}$ and $b\in\mathbb{R}^{MN}$ are weights and biases, and $f(.)$ is the activation function. When $W=diag\{\alpha,\alpha,\ldots,\alpha\}$, $b=\vec{0}$ and $f$ is the identity mapping $f(x)=x$, the MLP (\ref{mlp:brightness_mpl}) is equivalent to the brightness adjustment formula (\ref{con:brightness}). 
\\
\textbf{Contrast adjustment.} Contrast represents the difference in luminance or color maps. Among many definitions of contrast, we adopt a widely-used contrast adjustment formula:
\begin{equation} \label{con:contrast}
	I^{'}(x,y)=\alpha I(x,y)+(1-\alpha)\overline{I},
\end{equation}
where $\overline{I}=\frac{1}{M \times N} \sum_{x=0}^{M-1}\sum_{y=0}^{N-1} I(x,y)$ and $\alpha$ is the adjustment coefficient. When $\alpha=1$, the image will remain the same. The above formula is applied on each channel of the image. We can construct a three-layer MLP that is equivalent to the contrast adjustment operation. For simplicity, \textit{the following derivation is for a single-channel image,} and it can be easily generalized to RGB images (refer to the derivation of white-balancing in the supplementary material). As in Fig~\ref{fig:mlp}, the input layer has $M \times N$ units covering all pixels of the input image, the middle layer includes $M\times N+1$ hidden units and the last layer contains $M \times N$ output units. This can be formalized as:
\begin{equation}\label{mlp:contrast_mpl}
\begin{split}
Y=f_1(W_1^T X+b_1), Z=f_2(W_2^T Y+b_2)
\end{split}
\end{equation}
where $X\in\mathbb{R}^{MN}$,  $W_1\in\mathbb{R}^{MN \times (MN+1)}$,   $W_2\in\mathbb{R}^{(MN+1) \times MN}$, $b_1\in\mathbb{R}^{(MN+1)}$, $b_2\in\mathbb{R}^{MN}$. Let $A=diag\{\alpha,\alpha,\ldots,\alpha \}\in\mathbb{R}^{MN \times MN} $, $B=\frac{1}{MN}\vec{1}\in\mathbb{R}^{MN}$, $C=diag\{1,1,\ldots,1 \}\in\mathbb{R}^{MN \times MN}$, $D=[(1-\alpha)\vec{1}]^T\in\mathbb{R}^{1 \times MN}$. When $W_1=[A,B]\in\mathbb{R}^{MN \times (MN+1)}$, $W_2=
\begin{bmatrix}
C \\
D
\end{bmatrix}\in\mathbb{R}^{(MN+1) \times MN} $,
$b_1=b_2=\vec{0}$ and $f_1(x)=f_2(x)=x$,
the above MLP (\ref{mlp:contrast_mpl}) is equivalent to the contrast adjustment formula (\ref{con:contrast}).

Other operations, like white-balancing, saturation controlling, tone-mapping, can also be regarded as MLPs. (Please refer to the supplementary material.)\\
\textbf{Discussions.} We have shown that above the retouching operations are equivalent to classic MLPs. And the manipulation on one pixel is uncorrelated with neighboring pixels. That is why we can use a diagonal matrix as the MLP weights. Some operations, like contrast adjustment, also require global information (e.g., image mean value), which can be provided by another condition network. As shown in Figure~\ref{fig:mlp}(a) the above MLPs designed for input images can be viewed as MLPs worked on \textit{\textbf{individual pixels}}, which can be further formulated as $1\times1$ convolutions. The correlation between MLP and $1\times1$ convolutions has been revealed in MLPconv \cite{networkinnetwork} and SRCNN \cite{srcnn}. According to the analysis above, we propose a comprehensible and specialized framework for efficient photo retouching.

\subsection{Conditional Sequential Retouching Network}
The proposed framework contains a base network and a condition network as shown in Figure \ref{fig:mlp}(b). The base network takes the low-quality image as input and generates the retouched image. The condition network estimates the global priors from the input image, and afterwards influences the base network by global feature modulation operations.
\subsection*{3.2.1 Network Structure}
\textbf{Base network.} The base network adopts a fully convolutional structure with $N$ layers and $N-1$ ReLU activations. One unique trait of the base network is that all the filter size is $1\times1$, suggesting that each pixel in the input image is manipulated independently. Hence, the base network can be regarded as an MLP, which is worked \textit{on each pixel independently} and slides over the input image, as in \cite{networkinnetwork}. Based on the analysis in Section 3.1, theoretically, the base network has the capability of handling all the pixel-independent retouching operations. Moreover, since all the filters are of size $1\times1$, the network has dramatically few parameters.\\
\textbf{Condition network.} The global information/priors are indispensable for image retouching. For example, the contrast adjustment requires the average luminance of the image. To allow the base network to incorporate global priors, a condition network is proposed to collaborate with the base network. The condition network is like an encoder that contains three blocks, in which a series of convolutional, ReLU and downsamping layers are included. The output of the condition network is a condition vector, which will be broadcasted into the base network using the following global feature modulation. Network details are depicted in Section 4 and Figure~\ref{fig:mlp}(b).

\subsection*{3.2.2 Global Feature Modulation}
Our CSRNet adopts scaling and shifting operations to modulate intermediate features of the base network. First, we revisit the formulation of instance normalization \cite{instance}:
$
	IN(x_i) = \gamma * (\frac{x_i-\mu}{\sigma})  + \beta,
$
where $\mu$, $\beta$ are the mean and standard deviation of the feature map $x_i$, $\gamma$, $\beta$ are affine parameters.
The proposed Global Feature Modulation (GFM) only requires $\gamma$ and $\beta$ to scale and shift the feature map $x_i$ without normalizing it. Therefore, GFM can be formulated as:
$
	GFM(x_i) = \gamma * x_i + \beta.
$

GFM is also realted to Adaptive Feature Modification layer (AdaFM) \cite{adafm}, which can be written in the following equation:
$
	AdaFM(x_i) = g_i * x_i + b_i,
$
where $g_i$ and $b_i$ are the filter and bias. The modulation of AdaFM is based on a local region instead of a single pixel. In another perspective, GFM is a special case of AdaFM when the filter $g_i$ is of size $1\times1$. 

\begin{figure}[t]
	\centering
	\subfigure[View 1: Pixel-level]{
		\begin{minipage}[t]{\linewidth}
			\centering
			\includegraphics[width=4in]{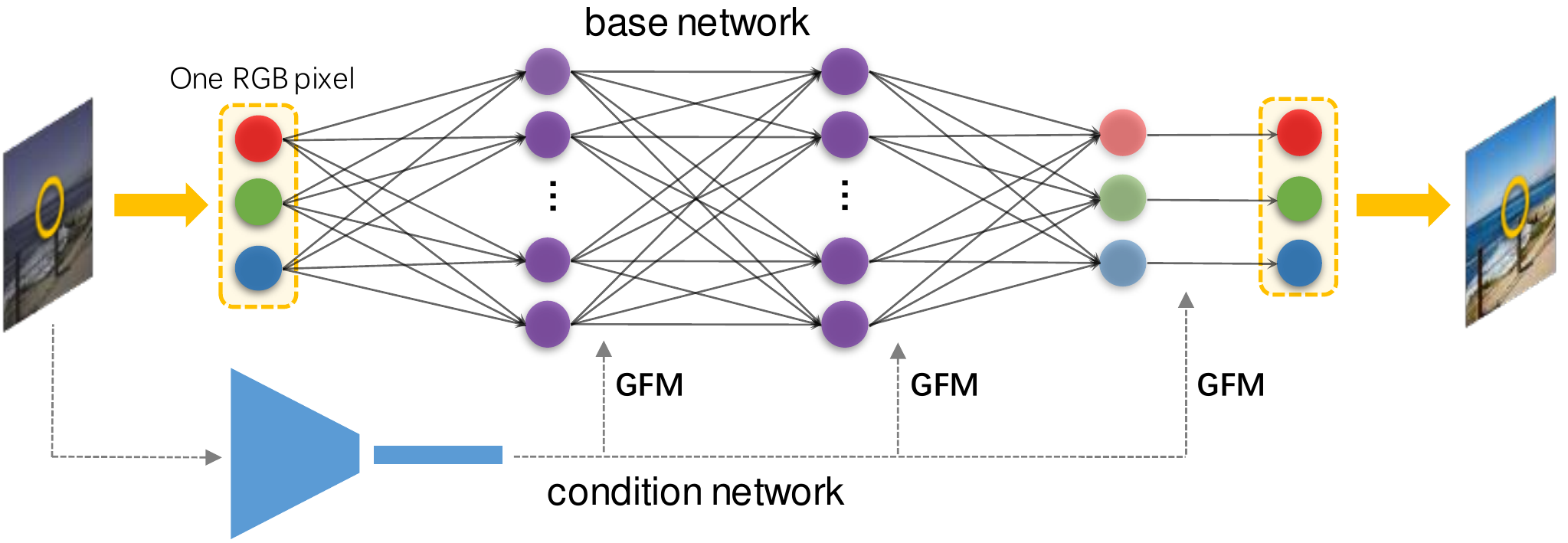}
		\end{minipage}
	}
	\subfigure[View 2: Space-level]{
		\begin{minipage}[t]{\linewidth}
			\centering
			\includegraphics[width=4in]{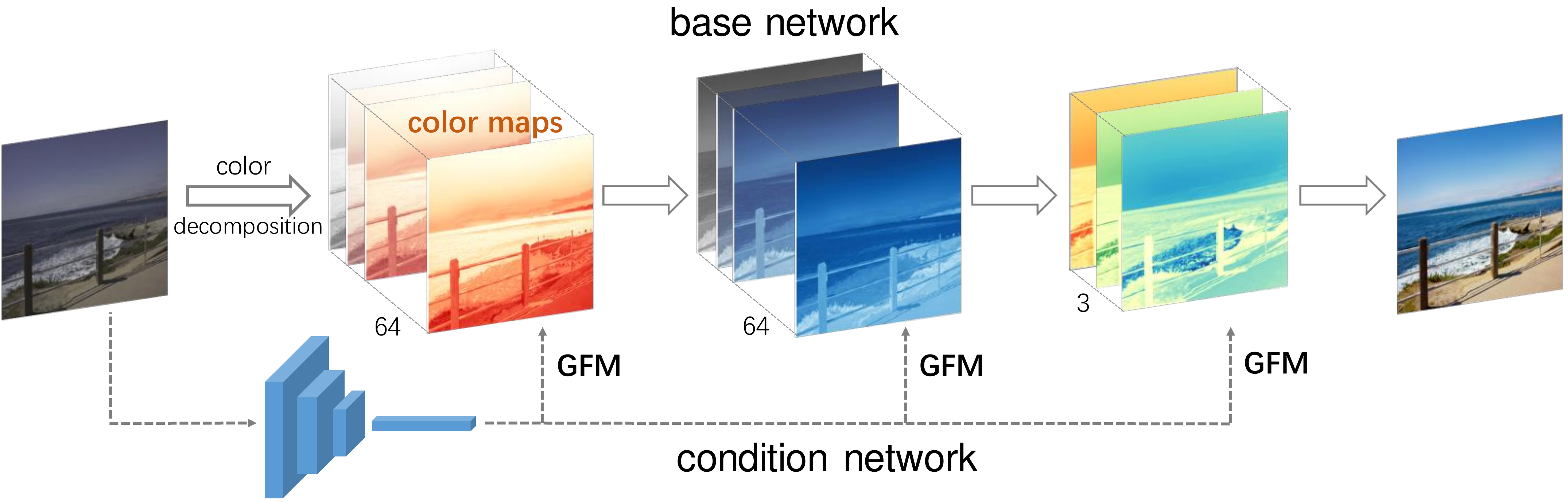}
		\end{minipage}
	}
	\caption{Illustration for two perspectives of the proposed framework.}
	\label{fig:views}
\end{figure}

\subsection*{3.2.3 Illustration}
To facilitate understanding, we illustrate “how the CSRNet works” in two perspectives. We use a simple yet standard setting --- $N=3$.\\
\textbf{Pixel-level view.} We regard the base network as an MLP that works on individual pixels as shown in Figure \ref{fig:views}(a). From this perspective, we can explain that the base network is made up of three fully-connected layers that perform feature extraction, non-linear mapping and reconstruction, respectively. As demonstrated in Section 3.1, such a three-layer MLP is able to approximate most image retouching operations. Then we add the condition network, and see how these two modules work collectively. As the GFM is equivalent to a ``multiply+addition" operation, it can be easily merged into filters. Then the condition network could adjust the filter weights of the base network. While for the last layer, the modulation operation can be modeled as an extra layer. This layer performs one-to-one mapping, which changes the “average intensity and dynamic range” of the output pixel, just like “brightness/contrast adjustment”. Combining the base network and the condition network, we will obtain a different MLP for a different input image, allowing image-specific photo retouching. 
To support this pixel-level view, we have conducted a demonstration experiment that using the proposed framework to simulate the procedures of several retouching operations. The results are shown in the supplementary material.\\

\textbf{Space-level view.} We can also regard intermediate features as color maps, while the color space transformation can be realized by linear combination of color maps (e.g., RGB to YUV).  Specifically, the input image is initialized in the RGB space. As depicted in Figure \ref{fig:views}(b), the first and second layers of the base network project the input into high dimensional color spaces, and the last layer transforms the color maps back to the RGB space. The GFM performs linear transformation on intermediate features, thus can be regarded as a retouching operation on the mapped color space. In summary, the base network performs color decomposition, the condition network generates editing parameters, and GFM sequentially adjusts  intermediate color maps.

\subsection*{3.2.4 Discussion}
In this part, we show the merits of CSRNet by comparing with other state-of-the-art methods. First, we adopt pixel-wise operations ($1\times1$ filters), which will preserve edges and textures. While GAN-based methods \cite{DPE,pix2pix} tend to change local patterns and generate undesired artifacts (see Figure \ref{fig:main}, Pix2Pix). Second, we use global modulation strategy, which will maintain color consistency across the image. While HDRNet \cite{hdrnet} predicts a transformation coefficient for each pixel, thus will lead to abrupt color changes (see Figure \ref{fig:main}, HDRNet). Third, we use a unified CNN framework with supervised learning, which could produce images of higher quality than RL-based methods \cite{white-box,Distort} (see Figure \ref{fig:main}, White-box and Distort-and-recover). Nevertheless, as CSRNet is specially designed for global photo retouching, it cannot be generalized to other tasks (e.g., style transfer, image enhancement, unpaired learning) as the above mentioned methods.

\subsection*{3.2.5 Multiple Styles and Strength Control}
Photo retouching is a highly ill-posed problem. Different photographers may have different preferences on retouching styles. Our method enjoys the flexibility of multiple style learning. Once the model is trained on a specific retouching style, we can easily transfer the model to another retouching style by only finetuning the condition network, which is much faster than training from scratch. While other methods all require to retrain the whole model on new datasets. Once we have obtained two stylized versions of an image, we can use ``image interpolation'' to produce intermediate styles between them.
\begin{equation} \label{fun:interpolation}
I_{out}(x,y)=\alpha \hat{I}_1(x,y) + (1-\alpha) \hat{I}_2(x,y),
\end{equation}
where $\hat{I}_1(x,y)$ and $\hat{I}_2(x,y)$ are two images to be interpolated, and $\alpha$ is the coefficient controlling the combined styles. Besides, this image interpolation strategy also allows us to control the overall retouching strength. For example, if the automatic retouched output is too bright, users may desire to decrease the overall luminance. This can be achieved by setting $I_1$ as the input image and $I_2$ as the retouched image. We can change the value of $\alpha$ to control the retouching strength. As shown in Figure \ref{fig:multi}, we could achieve continuous output effects between two objectives and two input images. There are other alternatives to realize continuous output effects, such as DNI \cite{DNI}, AdaFM \cite{adafm} and DynamicNet \cite{dynamic}. 
For pixel-wise operations, the pixels in two images on the same location are content-aligned, thus the pixel-wise blending is effective to achieve continuous imagery effects. If there are local operations, the blending should also consider neighboring pixels.
As photo retouching consists of only pixel-wise operations, the simplest image interpolation is already enough to achieve satisfactory results.

\section{Experiments}
\textbf{Dataset and Metrics.} MIT-Adobe FiveK \cite{mit-adobe} is a commonly-used photo retouching dataset with 5, 000 RAW images and corresponding retouched versions produced by five experts (A/B/C/D/E). We follow the previous methods \cite{white-box,DPE,Underexposed,hdrnet} to use the retouched results of expert C as the ground truth (GT). We adopt the same pre-processing procedure as \cite{white-box} \footnote{https://github.com/yuanming-hu/exposure/wiki/Preparing-data-for-the-MIT-Adobe-FiveK-Dataset-with-Lightroom} and all the images are resized to 500px on the long edge. We randomly select 500 images for testing and the remaining 4,500 images for training. We use PSNR, SSIM and the Mean L2 error in CIE L*a*b space \footnote{CIE L*a*b* (CIELAB) is a color space specified by the International Commission on Illumination. It describes all the colors visible to the human eye and was created to serve as a device-independent model to be used as a reference.} to evaluate the performance. \\
\textbf{Implementation Details.}
The base network contains 3 convolutional layers with channel size 64 and kernel size $1\times1$. 
The condition network also contains three convolutional layers with channel size 32. The kernel size of the first convolutional layer is set to $7\times7$ to increase the receptive field, while others are $3\times3$. Each convolutional layer downsamples features to half size with a stride of 2. We use a global average pooling layer at the end of the condition network to obtain a 32-dimensional condition vector. Then the condition vector will be transformed by fully connected layers to generate the parameters of channel-wise scaling and shifting operations. In total, there are 6 fully connected layers for 3 scaling operations and 3 shifting operations. During training, the mini-batch size is set to 1. L1 loss is adopted as the loss function. The learning rate is initialized as $10^{-4}$ and is decayed by a factor of 2 every $10^{5}$ iterations. All experiments run $6\times10^{5}$ iterations. We use PyTorch framework and train all models on GTX 2080Ti GPUs. It takes only 5 hours for the model training.

\subsection{Comparison with State-of-the-art Methods}
We compare our method with six state-of-the-art methods: DUPE \cite{Underexposed}, HDRNet \cite{hdrnet}, DPE \cite{DPE}, White-Box \cite{white-box}, Distort-and-Recover \cite{Distort} and Pix2Pix \cite{pix2pix}\footnote{Pix2Pix uses conditional generative adversarial networks to achieve image-to-image translation and is also applicable to image enhancement problem.}.

\textbf{Quantitative Comparison.}
We compare CSRNet with state-of-the-art methods\footnote{For White-Box, DUPE, DPE, we directly use their released pretrained models for testing. For HDRNet, Distort-and-Recover, and Pix2Pix, we re-train their models based on their public implementations on our training dataset. The training codes of DPE is not yet accessible and their released model is trained on another input version of MIT-Adobe FiveK. For fair comparison, we additionally train our models on the same input dataset.} in terms of PSNR, SSIM, and the Mean L2 error in L*a*b* space. As we can see from Table~\ref{table:main}, the proposed CSRNet outperforms all the previous state-of-the-art methods by a large margin with the fewest parameters (36,489). Specifically, White-Box and Distort-and-Recover are reinforcement-learning-based methods, which require over millions of parameters but achieve worst results. 
HDRNet and DUPE solve the color enhancement problem by estimating the illumination map and require relatively less parameters (less than one million). 

\begin{table}[htbp]
	\begin{center}
		\caption{Quantitative comparison with
			state-of-the-art methods on MIT-Adobe FiveK dataset (expert C). For L2 error in L*a*b space, lower is better.}
		\label{table:main}
		\begin{tabular}{lcccl}
			\hline\noalign{\smallskip}
			Method & PSNR & SSIM & L2 error (Lab) & params\\
			\noalign{\smallskip}
			\hline
			\noalign{\smallskip}
			White-Box \cite{white-box} & 18.59 & 0.797 & 13.24 & 8,561,762\\
			Distort-and-Recover \cite{Distort} & 19.54 & 0.800 & 12.91 & 259,263,320\\
			HDRNet \cite{hdrnet} & 22.65 & 0.880 & 11.64 & 482,080\\
			DUPE \cite{hdrnet} & 20.22 & 0.829 & 13.38 & 998,816\\
			Pix2Pix \cite{pix2pix} & 22.05 & 0.788 & 11.88 & 11,383,427\\
			\textbf{CSRNet (ours)} & \textbf{23.69} & \textbf{0.895} & \textbf{10.86} &36,489\\
			\hline
			\noalign{\smallskip}
			DPE \cite{DPE} & 23.76 & 0.881 & 10.50 & 3,335,395\\
			\textbf{CSRNet (ours)} & \textbf{24.23} & \textbf{0.900} & \textbf{10.29} & 36,489\\
			\hline
		\end{tabular}
	\end{center}
\end{table}

Since the released model of DUPE is trained for under-exposured images, we can also refer to the result (23.04dB) provided in their paper. Pix2Pix and DPE both utilize the generative adversarial networks and perform well quantitatively.
Under the same experimental setting, CSRNet outperforms DPE in all three metrics with much less parameters.

\begin{figure}[t]
	\centering
	\begin{minipage}[t]{0.24\textwidth}
		\centering
	\includegraphics[width=3cm,clip,trim=0cm 0.7cm 0cm 0.31cm]{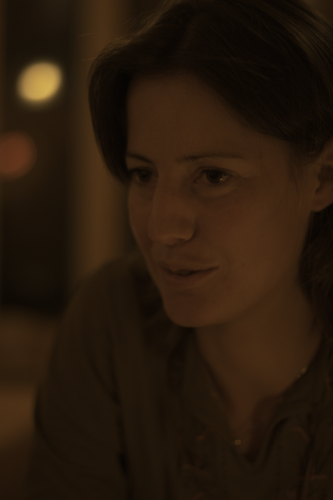}
		\scriptsize{input}
	\end{minipage}
	\begin{minipage}[t]{0.24\textwidth}
		\centering
		\includegraphics[width=3cm,clip,trim=0cm 5cm 0cm 2cm]{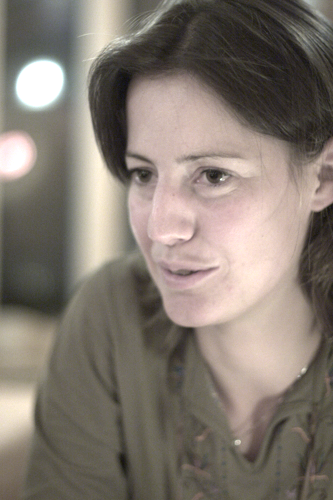}
		\scriptsize Distort-and-recover
	\end{minipage}
	\begin{minipage}[t]{0.24\textwidth}
		\centering
		\includegraphics[width=3cm,clip,trim=0cm 5cm 0cm 2cm]{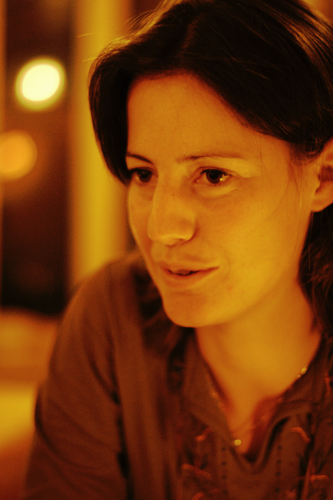}
		\scriptsize
		White-box
	\end{minipage}
	\begin{minipage}[t]{0.24\textwidth}
		\centering
		\includegraphics[width=3cm,clip,trim=0cm 5cm 0cm 2cm]{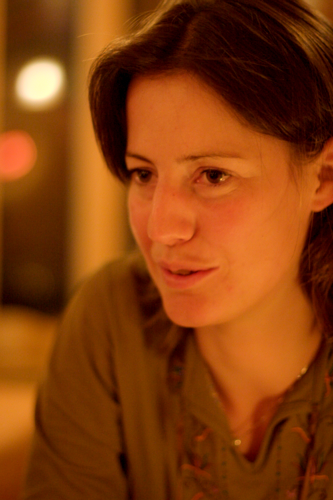}
		\scriptsize
		DPE
	\end{minipage}\\

	\begin{minipage}[t]{0.24\textwidth}
		\centering
		\includegraphics[width=3cm,clip,trim=0cm 5cm 0cm 2cm]{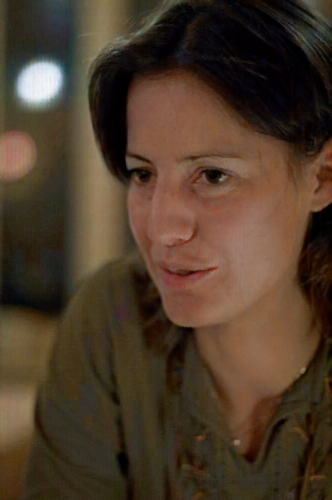}
		\scriptsize
		Pix2Pix
	\end{minipage}
	\begin{minipage}[t]{0.24\textwidth}
		\centering
		\includegraphics[width=3cm,clip,trim=0cm 5cm 0cm 2cm]{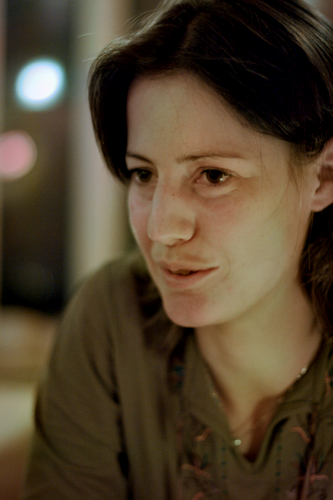}
		\scriptsize
		HDRNet
	\end{minipage}
	\begin{minipage}[t]{0.24\textwidth}
		\centering
		\includegraphics[width=3cm,clip,trim=0cm 5cm 0cm 2cm]{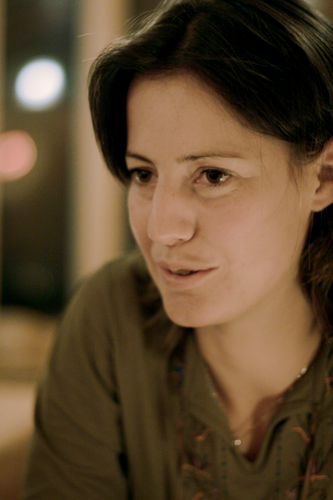}
		\scriptsize
		Ours
	\end{minipage}
	\begin{minipage}[t]{0.24\textwidth}
		\centering
		\includegraphics[width=3cm]{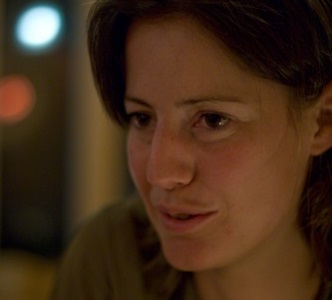}
		\scriptsize{GT}
	\end{minipage}\\

	\begin{minipage}[t]{0.24\textwidth}
		\centering
		\includegraphics[width=3cm]{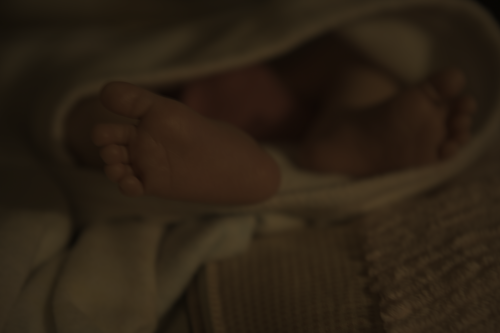}
		\scriptsize{input}
	\end{minipage}
	\begin{minipage}[t]{0.24\textwidth}
		\centering
		\includegraphics[width=3cm]{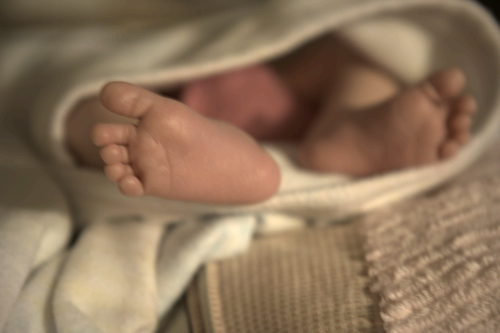}
		\scriptsize Distort-and-recover
	\end{minipage}
	\begin{minipage}[t]{0.24\textwidth}
		\centering
		\includegraphics[width=3cm]{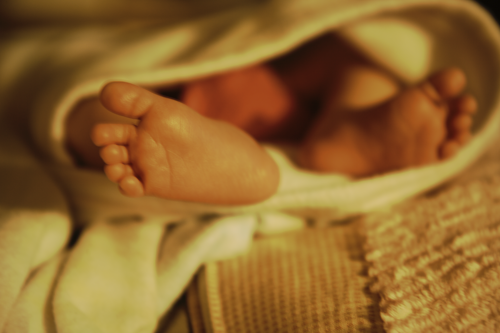}
		\scriptsize
		White-box
	\end{minipage}
	\begin{minipage}[t]{0.24\textwidth}
		\centering
		\includegraphics[width=3cm]{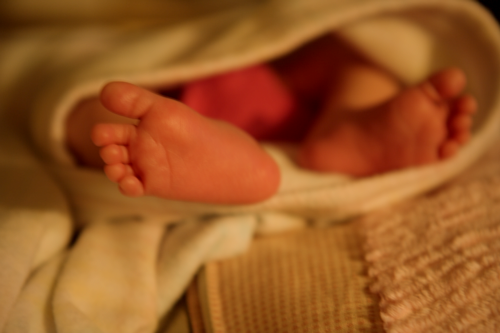}
		\scriptsize
		DPE
	\end{minipage}\\
	\begin{minipage}[t]{0.24\textwidth}
		\centering
		\includegraphics[width=3cm]{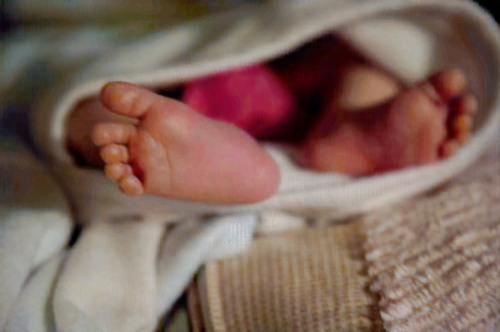}
		\scriptsize
		Pix2Pix
	\end{minipage}
	\begin{minipage}[t]{0.24\textwidth}
		\centering
		\includegraphics[width=3cm]{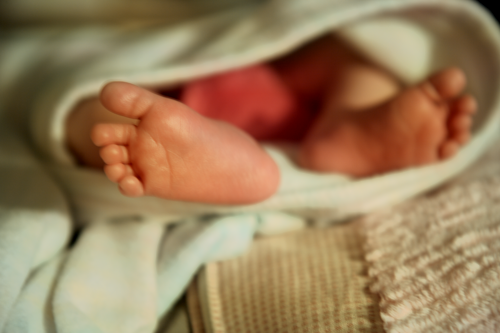}
		\scriptsize
		HDRNet
	\end{minipage}
	\begin{minipage}[t]{0.24\textwidth}
		\centering
		\includegraphics[width=3cm]{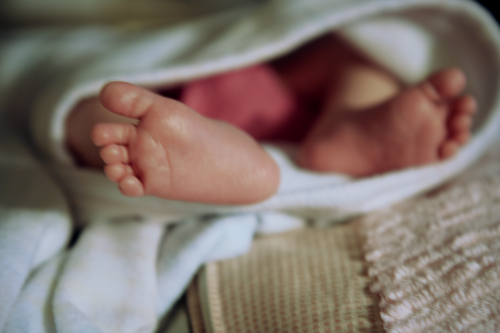}
		\scriptsize
		Ours
	\end{minipage}
	\begin{minipage}[t]{0.24\textwidth}
		\centering
		\includegraphics[width=3cm]{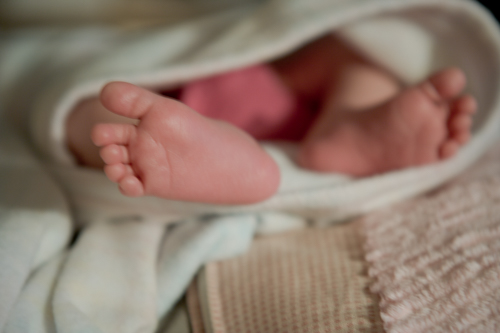}
		\scriptsize
		GT
	\end{minipage}
	\caption{Visual comparison with state-of-the-arts on MIT-Adobe FiveK data set.}
	\label{fig:main}
\end{figure}

\textbf{Visual Comparison.}
The results of visual comparison\footnote{We do not consider DUPE for visual comparison because the authors only released model trained on their collected under-exposured image pairs.} are shown in Figure \ref{fig:main}. The input images from the MIT-Adobe FiveK dataset are generally under low-light condition. Distort-and-recover tends to generate over-exposure output. It seems that White-box and DPE only increase the brightness but fail to modify the original tone, which is oversaturated. At the first glance, the outputs of the second row look more natural and vivid. However, the enhanced image obtained by Pix2Pix contains artifacts. HDRNet outputs image with unnatural color in some regions (e.g. green color on the face). In conclusion, our method is able to generate more realistic images among all methods. Please see the supplementary file for more comparisons.

\textbf{User Study.}
We have conducted a user study with 20 participants for subjective evaluation. The participants are asked to rank four retouched image versions (HDRNet \cite{hdrnet}, DPE \cite{DPE}, expert-C (GT) and ours) according to the aesthetic visual quality. 50 images are randomly selected from the testing set and are shown to each participant. 4 retouched versions are displayed on the screen in random order. Users are asked to pay attention to whether the color is vivid,whether there are artifacts and whether the local color is harmonious. Since HDRNet and DPE have better quantitative and qualitative performance than other methods, we choose them to make the comparison. As suggested in Figure \ref{fig:user_study}, our results achieve better visual ranking against HDRNet and DPE with 553 images ranked first and second. 245 images of our method ranked first, second only to expert C; and 308 images are ranked second, ahead of other methods.

\begin{minipage}[t]{\textwidth}
	\flushleft
	\begin{minipage}[t]{0.48\textwidth}			
		\makeatletter\def\@captype{figure}\makeatother\caption{Ranking results of user study. Rank 1 means the best visual quality.}\label{fig:user_study}	
		\includegraphics[width=2.1in]{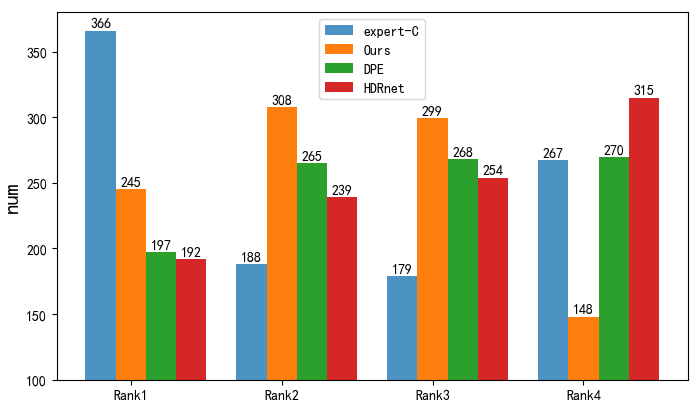}
	\end{minipage}
	\begin{minipage}[t]{0.48\textwidth}
		\centering
		\makeatletter\def\@captype{table}\makeatother\caption{Performance for Multiple styles (A/B/D/E).}
		\label{table:multi}
		\renewcommand{\arraystretch}{1}
		\setlength{\tabcolsep}{3pt}
		\begin{tabular}{ccc}
			\hline\noalign{\smallskip}
			expert & \makecell{PSNR \\ (finetune)} &  \makecell{PSNR \\ (scratch)} \\
			\noalign{\smallskip}
			\hline
			\noalign{\smallskip}
			A & 22.29 & 22.06\\
			B & 25.61 & 25.52\\
			D & 23.06 & 23.04 \\
			E & 23.95 & 23.81 \\
			\hline
		\end{tabular}
	\end{minipage}
\end{minipage}

\subsection{Multiple Styles and Strength Control}
In this section, we aim to achieve different retouching styles and control retouching strength. Specifically, given an image retouching model for one style, we can easily transfer the model to other retouching styles by only finetuning the condition network. Here, we transfer the retouching model of expert C to expert A, B, D, and E. From Table~\ref{table:multi}, we can observe that finetuning the condition network can achieve comparable results with training from scratch. This indicates that the fixed base network performs like a stacked color decomposition, and have the flexibility to be modulated to different retouching styles.

\begin{figure}[t]
	\centering
	\centering
	\begin{minipage}[t]{0.19\textwidth}
		\centering
		\includegraphics[width=2.4cm]{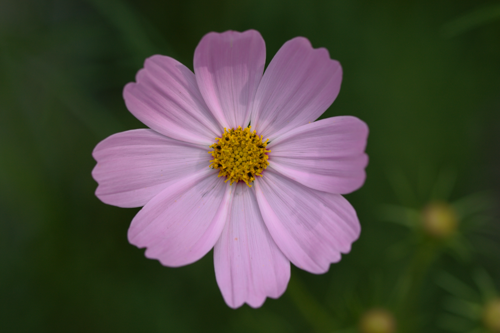}
		\scriptsize
		\textbf{expert A} $\alpha=0.0$
	\end{minipage}
	\begin{minipage}[t]{0.19\textwidth}
		\centering
		\includegraphics[width=2.4cm]{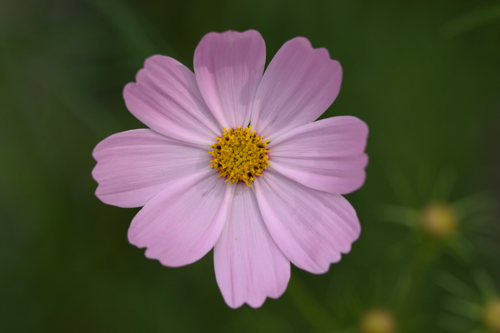}
		\scriptsize
		$\alpha=0.3$
	\end{minipage}
	\begin{minipage}[t]{0.19\textwidth}
		\centering
		\includegraphics[width=2.4cm]{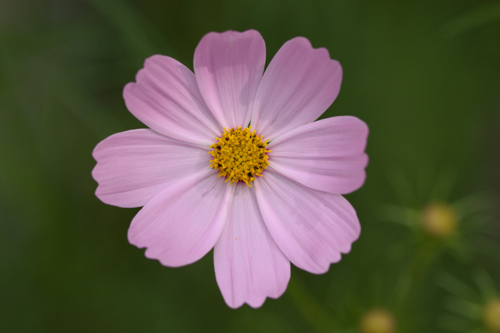}
		\scriptsize
		$\alpha=0.5$
	\end{minipage}
	\begin{minipage}[t]{0.19\textwidth}
		\centering
		\includegraphics[width=2.4cm]{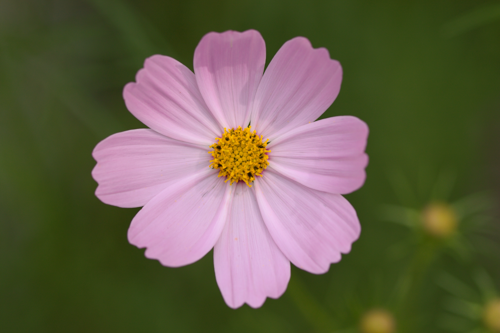}
		\scriptsize
		$\alpha=0.8$
	\end{minipage}
	\begin{minipage}[t]{0.19\textwidth}
		\centering
		\includegraphics[width=2.4cm]{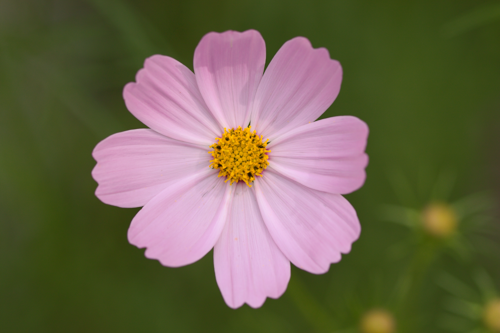}
		\scriptsize
		$\alpha=1.0$ \textbf{expert B}
	\end{minipage}\\
\begin{minipage}[t]{0.19\textwidth}
	\centering
	\includegraphics[width=2.4cm]{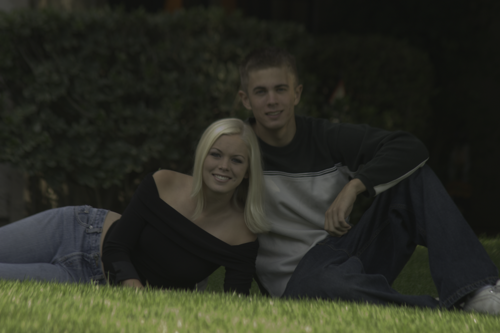}
	\scriptsize
	\textbf{input} $\alpha=0.0$
\end{minipage}
\begin{minipage}[t]{0.19\textwidth}
	\centering
	\includegraphics[width=2.4cm]{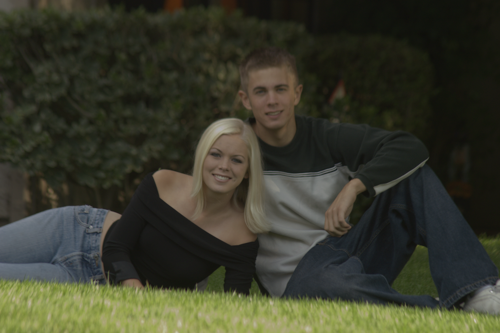}
	\scriptsize
	$\alpha=0.3$
\end{minipage}
\begin{minipage}[t]{0.19\textwidth}
	\centering
	\includegraphics[width=2.4cm]{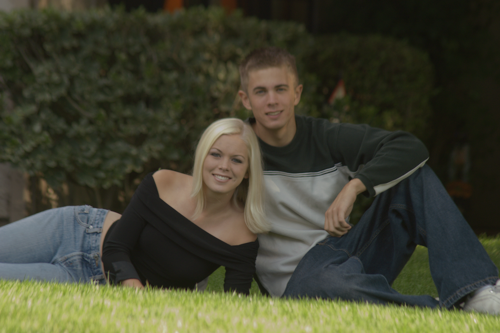}
	\scriptsize
	$\alpha=0.5$
\end{minipage}
\begin{minipage}[t]{0.19\textwidth}
	\centering
	\includegraphics[width=2.4cm]{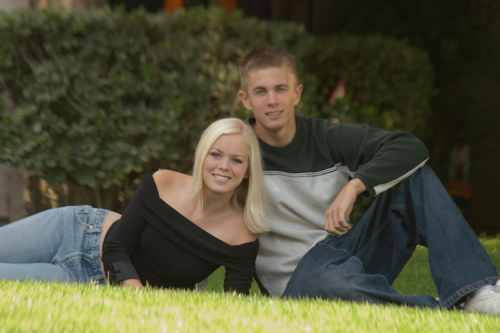}
	\scriptsize
	$\alpha=0.8$
\end{minipage}
\begin{minipage}[t]{0.19\textwidth}
	\centering
	\includegraphics[width=2.4cm]{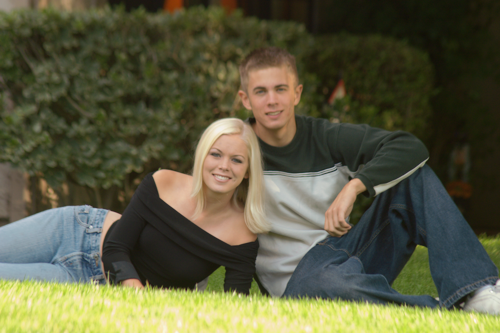}
	\scriptsize
	$\alpha=1.0$ \textbf{expert B}
\end{minipage}\\
	\caption{The first row shows smooth transition effects between different styles (expert A to B) by image interpolation. In the second row, we use image interpolation to control the retouching strength from input image to the automatic retouched result. We denote the interpolation coefficient $\alpha$ for each image.} 
	\label{fig:multi}
\end{figure}

Given retouched outputs of different styles, users can achieve smooth transition effects between different styles by using image interpolation. In Figure~\ref{fig:multi}, the output style changes continuously from expert A to expert B. 
Besides, for one certain style, users can also control the retouching strength by image interpolation between input image and the retouched one. More results can be found in the supplementary file.

\subsection{Ablation study}
In this section, we investigate our CSRNet in three aspects, base network, modulation strategy, and condition network. We present all the results in PSNR.

\textbf{Base network.}
The base network of our CSRNet contains 3 convolutional layers with kernel size $1\times1$ and channel number $64$.
As mentioned before, we assume that the base network with kernel size $1\times1$ performs like a stacked color decomposition, and each layer represents a different color space of the input image. 
Here, we explore the base network by changing its kernel size and increasing the number of layers. Besides, we remove the condition network to verify whether the base network could fairly deal with image retouching alone.

\begin{minipage}{\textwidth}
	\begin{center}
	\makeatletter\def\@captype{table}\makeatother\caption{Results of ablation study for the base network.}
	\label{table:base_net}
		\small
		\renewcommand{\arraystretch}{0.8}
		\setlength{\tabcolsep}{3pt}
		\begin{tabular}{ccccl}
			\hline\noalign{\smallskip}
			&layers&kernel size& PSNR  & params\\
			\noalign{\smallskip}
			\hline
			\noalign{\smallskip}
			w/o condition &3& $1\times1$ & 20.47 & 4,611 \\
			&3& $3\times3$  & 20.69 & 40,451 \\
			&7& $3\times3$  & 20.67 & 188,163\\
			\hline
			\noalign{\smallskip}
			w condition &3& $1\times1$   & 23.69 &  36,489 (ours)\\
			&3& $3\times3$  & 23.73 &  72,329\\
			&5& $1\times1$  & 23.73 &  53,257\\
			&5& $3\times3$  & 23.70 &  154,633\\
			&7& $1\times1$  & 23.83 &  70,025\\
			&7& $3\times3$  & 23.64 &  236,937\\
			\hline
		\end{tabular}
	\end{center}
\end{minipage}

From Table~\ref{table:base_net}, we can observe that the base network cannot solve the image retouching problem well without the condition network. Specifically, when we expand the filter size to $3\times3$ and increase the number of layers to 7, there is only marginal improvement (0.2dB) in terms of PSNR. 

Considering the cases with condition network, if we fix the number of layers, and expand the kernel size to $3\time3$, there is roughly no improvement. Therefore, the sequential processing of the base network is just pixel-independently which can be achieved by $1\times1$ filters. 
If we fix the kernel size to $1\times1$ and increase the number of layers, the performance improves a little bit (0.14dB). 
Since more layers require more parameters, we adopt a light-weight architecture with only three layers.

\textbf{Modulation strategy.}
Our framework adopts GFM to modulate the intermediate features under different conditions. Here, we compare different modulation strategies: concatenating, AdaFM \cite{adafm}, and SFTNet \cite{sftnet}. Specifically, we concatenate the condition vector directly with the input image. For AdaFM, we use kernel size $3\times3$ and $5\times5$. For SFTNet, we remove all the stride operations and the global average pooling in the condition network. Therefore, the modified condition network is able to generate a condition map, thus allowing spacial feature modulation on the intermediate features.

From Table~\ref{table:modulation}, we observe that SFTNet obtains the worst results compared with other modulation strategies. Therefore, image retouching mainly depends on global context rather than spacial information. As for AdaFM, it is hard to achieve improvement by simply expanding its kernel size. In conclusion, conditional image retouching can be effectively achieved by GFM, which only scales and shifts the intermediate features.

\begin{minipage}{\textwidth}
	\begin{minipage}[t]{0.36\textwidth}
		\scriptsize
		\makeatletter\def\@captype{table}\makeatother\caption{Ablation study on modulation strategy.}
		\label{table:modulation}
		\renewcommand{\arraystretch}{1.22}
		\setlength{\tabcolsep}{4pt}
		\begin{tabular}{lcl}
			\hline\noalign{\smallskip}
			\makecell{modulation \\ strategies}  & PSNR  & params\\
			\noalign{\smallskip}
			\hline
			\noalign{\smallskip}
			w/o condition  & 20.47 & 4,611 \\
			concat & 23.31 & 29,891\\
			AdaFM $3\times3$ & 23.70 & 71,073 \\
			AdaFM $5\times5$  & 23.38 & 140,241 \\
			SFTNet & 20.73  & 36,489\\
			CSRNet & 23.69 & 36,489 \\
			\hline
		\end{tabular}
	\end{minipage}
	\begin{minipage}[t]{0.6\textwidth}
		\scriptsize
		\centering
		\makeatletter\def\@captype{table}\makeatother\caption{Results of ablation study for the condition network}
		\label{table:condition}
		\renewcommand{\arraystretch}{0.96}
		\setlength{\tabcolsep}{2pt}
		\begin{tabular}{lllcl}
			\hline\noalign{\smallskip}
			&global prior & dim & PSNR & params  \\
			\noalign{\smallskip}
			\hline
			\noalign{\smallskip}
			w/o condition& None & 0 & 20.47 & 4,611 \\
			\quad \; \;network&brightness & 1 & 21.47 & 5,135\\
			&average intensity & 3 & 21.93 & 5,659\\
			&histograms & 768 & 22.90 & 206,089\\
			\noalign{\smallskip}
			\hline
			\noalign{\smallskip}
			w condition& None (ours)& 32 & 23.69 & 36,489\\
			\quad network&brightness & 1$+$32 & 23.01 & 36,751\\
			&average intensity & 3$+$32 & 23.57& 37,275\\
			&histograms & 768$+$32 & 23.39 & 237,705\\
			
			\hline
		\end{tabular}
	\end{minipage}
\end{minipage}

\textbf{Condition network.}
The condition network aims to estimate a condition vector that represents global information of the input image. Alternatively, we can use other hand-crafted global priors to control the base network, such as brightness, average intensity, and histograms. Here, we investigate the effectiveness of these global priors. 
For brightness, we transform the RGB image to gray image, while the mean value of the gray image is regarded as the global prior. For average intensity, we compute the mean value for each channel of the RGB image. Regarding histograms, we generate the histograms for each channel of RGB image, and then concatenate them to a single vector.
Besides, we combine the global priors with our condition network to control the base network. In particular, we concatenate the global prior with the condition vector produced by the condition network. In addition, we have also explored the condition network with different hyper-parameters. The experimental results can be found in the supplementary file.

From Table~\ref{table:condition}, all three global priors can largely improve the performance compared with base network alone, which means that global priors are essential for image retouching. The ranking of their effectiveness is: histograms $\textgreater$ average intensity $\textgreater$ brightness. 
However, it seems that simply concatenating the global prior with condition vector cannot achieve improvement. In conclusion, our CSRNet can already extract effective global information.


\section{Acknowledgement}
This work is partially supported by the National Natural Science Foundation of China (61906184), Science and Technology Service Network Initiative of Chinese Academy of Sciences (KFJ-STS-QYZX-092), Shenzhen Basic Research Program (JSGG20180507182100698, CXB201104220032A), the Joint Lab of CAS-HK，Shenzhen Institute of Artificial Intelligence and Robotics for Society.

\bibliographystyle{splncs04}

\end{document}